\documentclass{article}

\usepackage[final]{neurips_2025}

\usepackage[utf8]{inputenc} 
\usepackage[T1]{fontenc}    
\usepackage{hyperref}       
\usepackage{url}            
\usepackage{booktabs}       
\usepackage{amsfonts}       
\usepackage{nicefrac}       
\usepackage{microtype}      
\usepackage{xcolor}         
\usepackage{mathtools}
\usepackage{times}
\usepackage{latexsym}
\usepackage{amsmath}

\usepackage[T1]{fontenc}
\usepackage[utf8]{inputenc}
\usepackage{microtype}
\usepackage{inconsolata}
\usepackage{graphicx}
\usepackage{multirow}

\usepackage{pifont}
\usepackage{amsmath}
\usepackage{amssymb}
\usepackage[marginal]{footmisc}
\usepackage{wrapfig} 
\usepackage{graphicx}
\usepackage[caption=false]{subfig}
\usepackage{booktabs}
\usepackage{tabularx}

\title{SeniorTalk: A Chinese Conversation Dataset with Rich Annotations for Super-Aged Seniors}

\author{%
\textbf{Yang Chen}$^1$\thanks{Equal Contribution. $\dag$ Corresponding author.} \And \textbf{Hui Wang}$^{1*}$ \And \textbf{Shiyao Wang}\textsuperscript{1} \And \textbf{Junyang Chen}\textsuperscript{1}\And \textbf{Jiabei He}\textsuperscript{1} \And
 \textbf{Jiaming Zhou}\textsuperscript{1} \And \textbf{Xi Yang}\textsuperscript{2}\And \textbf{Yequan Wang}\textsuperscript{2}\And \textbf{Yonghua Lin}\textsuperscript{2}\And \textbf{Yong Qin}$^{1 \dag}$
\\
 \textsuperscript{1}College of Computer Science, Nankai University,
 \\
 \textsuperscript{2}Beijing Academy of Artificial Intelligence, Beijing, China,
}

\begin{document}
\maketitle

\begin{abstract}
While voice technologies increasingly serve aging populations, current systems exhibit significant performance gaps due to inadequate training data capturing elderly-specific vocal characteristics like presbyphonia and dialectal variations. The limited data available on super-aged individuals in existing elderly speech datasets, coupled with overly simple recording styles and annotation dimensions, exacerbates this issue. To address the critical scarcity of speech data from individuals aged 75 and above, we introduce SeniorTalk, a carefully annotated Chinese spoken dialogue dataset. This dataset contains 55.53 hours of speech from 101 natural conversations involving 202 participants, ensuring a strategic balance across gender, region, and age. Through detailed annotation across multiple dimensions, it can support a wide range of speech tasks. We perform extensive experiments on speaker verification, speaker diarization, speech recognition, and speech editing tasks, offering crucial insights for the development of speech technologies targeting this age group. Code is available at \url{https://github.com/flageval-baai/SeniorTalk}  and data at \url{https://huggingface.co/datasets/evan0617/seniortalk}.





\end{abstract}

\section{Introduction}

The rapid global aging population presents both significant challenges and opportunities in the development of technologies specifically designed for older adults, particularly those aged 75 and over \citep{chumuang2024voice,SCUTERI202435}. As the number of people in this ultra-high-age group continues to grow, it becomes increasingly important to enhance the accessibility and inclusion of speech-based technologies for this demographic \citep{ portet2013design, young2010difficulties}. However, many state-of-the-art speech systems struggle to perform effectively within the elderly population, exhibiting biases on elderly vocal patterns. \cite{kulkarni-etal-2024-balancing, feng2021quantifying}. For instance, existing speech recognition systems often show poor performance with older users, influenced by factors such as speech deterioration, hearing loss, health issues \citep{fraser2015linguistic}, and the diversity of speech patterns among older adults \citep{geng2022speaker, 10584335}. A key underlying reason for this challenge is the lack of datasets specifically addressing the unique needs of ultra-high-age individuals \cite{10584335}, which hampers the development of robust foundation models and the application of tailored solutions.

Although existing studies have made some efforts to collect geriatric speech data \citep{sekerina2024brazilian, fukuda-etal-2020-improving}, significant limitations remain. First, current speech corpora predominantly focus on younger adults or healthy senior populations, with samples that consist mainly of scripted speech and exhibit standardized accents. 
For example, both the AISHELL-ASR0060 and MECSD datasets \citep{wang19q_interspeech} adopt an unusually low age threshold of 55 years for defining elderly participants. In particular, these data sets show extremely low proportions of participants aged 75 to 85 years: only 0.8\% for AISHELL-ASR0060 and 9.4\% for MECSD, which means that the oldest population (75+ years) is severely underrepresented. Similarly, the S-JNAS corpus of elderly Japanese speech reports a mean speaker age of 67.6 years \citep{fukuda-etal-2020-improving}. These fall below the World Health Organization's (WHO) geriatric classification designating 75+ years as the late elderly phase which is a period associated with progressive age-related decline in physiological function \citep{orimo2006reviewing}.

Second, the current collection paradigms and annotation methods of existing datasets further limit their practical applicability. A large portion of speech resources for the elderly primarily focuses on reading style \citep{wang19q_interspeech,7357859}, which do not reflect the everyday communication scenarios that elderly people encounter in real life. Furthermore, many corpora are tailored for narrowly defined tasks such as automatic speech recognition (ASR) or pathology detection \citep{wang19q_interspeech}, thereby preventing comprehensive characterization of age-related vocal variations. These datasets also lack key features such as speaker diarization or dialect labels, which restrict their ability to support a wider array of use cases and hinder their robustness in addressing the diverse challenges of speech processing.

To address these limitations, we introduce \textbf{SeniorTalk}, the only freely available open-source Mandarin speech dataset consisting of spontaneous conversations among individuals aged 75 and older. As shown in Table~\ref{tab:chinese}, this dataset comprises natural conversational recordings from 202 native Chinese speakers, representing a rich diversity of regional, age, and gender demographics, and captured in authentic, real-world interaction settings. It effectively addresses the current gaps and limitations in datasets focused on elderly populations, particularly the underrepresentation of super-aged seniors and the lack of diversity in recording styles and annotation dimensions. Moreover, we conduct extensive experiments across various speech tasks, providing a benchmark specifically for the elderly population. By open-sourcing this corpus along with fine-grained metadata, we aim to bridge the vocal age gap and promote the development of equitable voice technologies for aging societies. SeniorTalk makes three key contributions to the field:
\begin{itemize}

    \item We recruit 202 super-elderly speakers from 16 provinces in China, ensuring balance across gender, age, and geography. This resulted in SeniorTalk, a dataset with 55.53 hours of data from 101 conversations.

    \item We provide detailed, multi-dimensional annotations for the dataset, encompassing speaker information, transcriptions, timestamps, and more. These annotations enable comprehensive speech signal analysis and support a wide range of speech-related tasks for the elderly.
 
    \item We conduct a comprehensive series of experiments using the dataset, covering tasks such as speaker verification, speaker diarization, speech recognition, and others. This establishes a solid benchmark for evaluating models across various speech-related tasks.
    
\end{itemize}

\begin{table}
\caption{Comparison of Chinese elderly speech datasets, including AISHELL-ASR0060 (marked in the table as ASR0060), MECSD, and SeniorTalk, across annotation features.}
\centering
\resizebox{\textwidth}{!}{
\begin{tabular}{@{}ccccccccccc@{}}
\toprule
\multirow{2}{*}{Dataset} & \multirow{2}{*}{\shortstack{Mean \\ Age}} & \multirow{2}{*}{Style} & \multicolumn{5}{c}{Annotation Features}  & \multicolumn{3}{c}{Age Group Distribution (\%)} \\
\cmidrule(lr){4-8}\cmidrule(lr){9-11}
 & & &  Region & Transcript & Timestamp & Accent  & Sound  & 55-65 & 65-75 & 75-85 \\
\midrule
ASR0060 & 60.42 & Reading & North/South/Other & $\checkmark$  & N/A & $\times$ & $\times$  & 56.7 & 42.5 & 0.8 \\
MECSD & 67.27  & Reading & $\times$ & $\checkmark$ & N/A & $\times$ & $\checkmark$  & 17.6 & 72.9 & 9.4 \\
SeniorTalk & 79.5 & Conversation & Provincial & $\checkmark$ & $\checkmark$ & $\checkmark$ & $\checkmark$& 0 & 0 & 100 \\
\bottomrule
\end{tabular}
}
\label{tab:chinese}
\end{table}


\section{Related Work}

\begin{table}[t]
\caption{Summary of related elderly speech datasets. Key characteristics include the age ranges of speakers (Age), the number of speakers (\# Spks.), publication year (Year), and availability status (Avail.), where 'P' indicates partial availability.}
\centering
\begin{tabular}{p{3cm}p{1.5cm}cp{1.5cm}cccc}
\toprule
\textbf{Corpus}              & \textbf{Language} & \textbf{Age} & \textbf{Style }& \# \textbf{Spks.} & \textbf{Dur.(hrs)} & \textbf{Year} & \textbf{Avail.}\\
\midrule
BraPoRus \citep{sekerina2024brazilian} & Brazilian Portuguese-Russian & 59-98 &Monologue, interview, ...& 1,500 &170 & 2024 & N\\
EARS \citep{FUKUDA2023101424}& Japanese & 70-99 &Reading &123 &13.4 &2023& N\\

Improving S-JNAS \citep{fukuda-etal-2020-improving}& Japanese&  65-99 &Reading & 221 & 31.7 & 2020 & Y \\
AISHELL-ASR0060   & Mandarin &  55+ &Reading & 503  & 793  & 2019  & Y\\
MECSD \citep{wang19q_interspeech} & Mandarin  &  55-85 &Reading & 85  & 110  & 2019 & P\\
elderLUCID \citep{hazanelderlucid} & English & 19-84&Reading & 83 & N & 2017 & N\\
Develop S-JNAS \citep{7357859}& Japanese&  60-98 &Reading & 100 & 9.2 & 2015 & Y \\
ERES38 \citep{aman2013speech} & French &  68-98 &Interview & 22  & 17  & 2013 & N \\
AD80 \citep{aman2013speech} & French &  62-94 &- & 43   &  4.7  & 2013 & N \\
CCC \citep{PopeDavis+2011+143+161} & English &  65+ &Interview & 600+  & 800+  & 2011 & Y \\
S-JNAS & Japanese&  60-90 &Reading & 301 & - & 2007 & Y \\
E-MIC & Korean&  65-85 &conversation & 100 & 3 & - & Y \\
\bottomrule
\end{tabular}

\label{dataset_survey}
\end{table}


Several corpora have been developed to address the specific acoustic characteristics of elderly speech. Early work in this area includes the Japanese Newspaper Article Sentences Read Speech Corpus of the Aged (S-JNAS), a foundational resource for Japanese elderly speech research. Subsequent research has broadened the scope of investigation, with efforts such as the Carolinas Conversations Collection (CCC) \citep{PopeDavis+2011+143+161} focusing on multiethnic elderly speakers with chronic conditions, offering valuable insights into how sociolinguistic factors and health status influence speech production. The AD80 and ERES38 corpora \citep{aman2013speech} advance French elderly speech analysis through distress detection benchmarks for ambient assisted living systems.  Further corpus development has continued with the creation of specialized datasets, including a corpus of 100 elderly Japanese speakers designed to enhance human-robot interaction in elder care \citep{7357859}.  The elderLUCID project \citep{hazanelderlucid} examines the complexities of speech communication in older adults, considering the interplay of hearing loss, phonation, articulation, and cognitive factors.  

More recently, researchers have focused on specific languages and conditions, as exemplified by the Mandarin Speech Database for Early Dementia Detection (MECSD) \citep{wang19q_interspeech} and the AISHELL-ASR0060 database\footnote{\url{https://www.aishelltech.com/General_Datasets}} for elderly Mandarin speech.  Improvements to existing corpora, such as S-JNAS, have also been explored, including the creation of acoustic models specifically designed for "super-elderly" speakers \citep{fukuda-etal-2020-improving}.  The Elderly Multimodal Interpersonal Conversation (E-MIC)\footnote{\url{https://ai4robot.github.io/emic-en/}} dataset expands the scope of analysis by incorporating multimodal data, including video and audio, to study turn-taking in elderly conversations.  Further work, such as EARS \citep{FUKUDA2023101424}, has continued to refine acoustic modeling techniques for super-elderly Japanese speakers, while the BraPoRus corpus \citep{sekerina2024brazilian} highlights the importance of preserving heritage languages and the challenges of remote data collection during the COVID-19 pandemic.

\section{Dataset description}
\label{sec:2}

\subsection{Dataset overview}

SeniorTalk is designed specifically for the ultra-elderly population aged 75 and above, offering a comprehensive collection of spoken dialogue data aimed at supporting various speech-related tasks. The dataset includes a total of 101 recorded speech dialogues, representing a diverse range of linguistic characteristics. It spans 55.53 hours of speech data, recorded from 202 participants, and features 60,029 individual utterances. Additionally, the dataset is enriched with annotations across 8 distinct dimensions, ensuring its suitability for training and evaluating robust models across multiple speech-processing tasks.


\begin{figure}[t]
  \centering
  \subfloat[Population by age and gender]{\includegraphics[width=0.48\columnwidth]{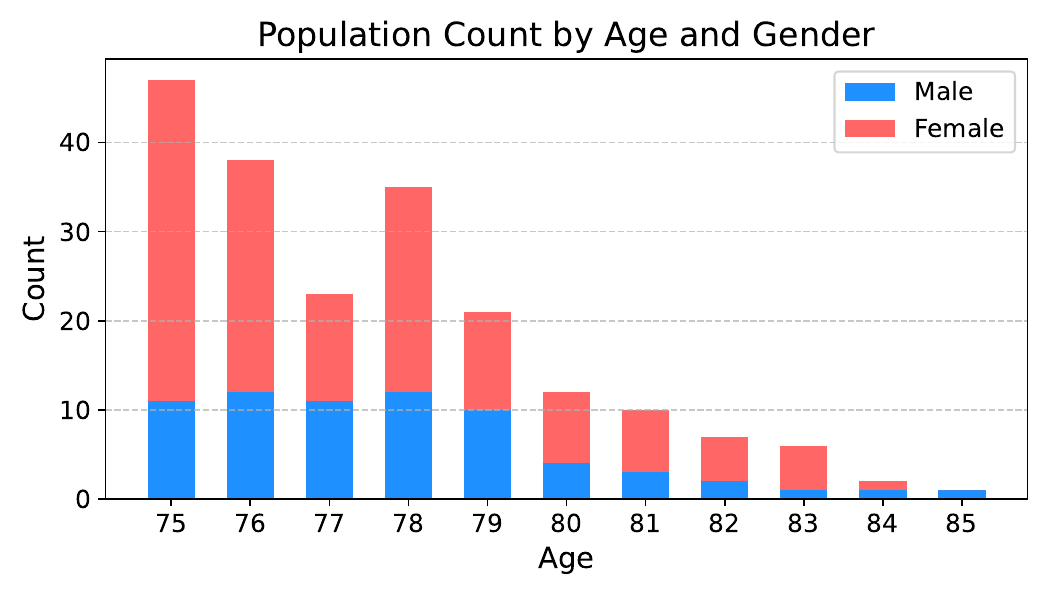}\label{fig:agea}}
  \hfill
  \subfloat[Duration distribution]{\includegraphics[width=0.48\columnwidth]{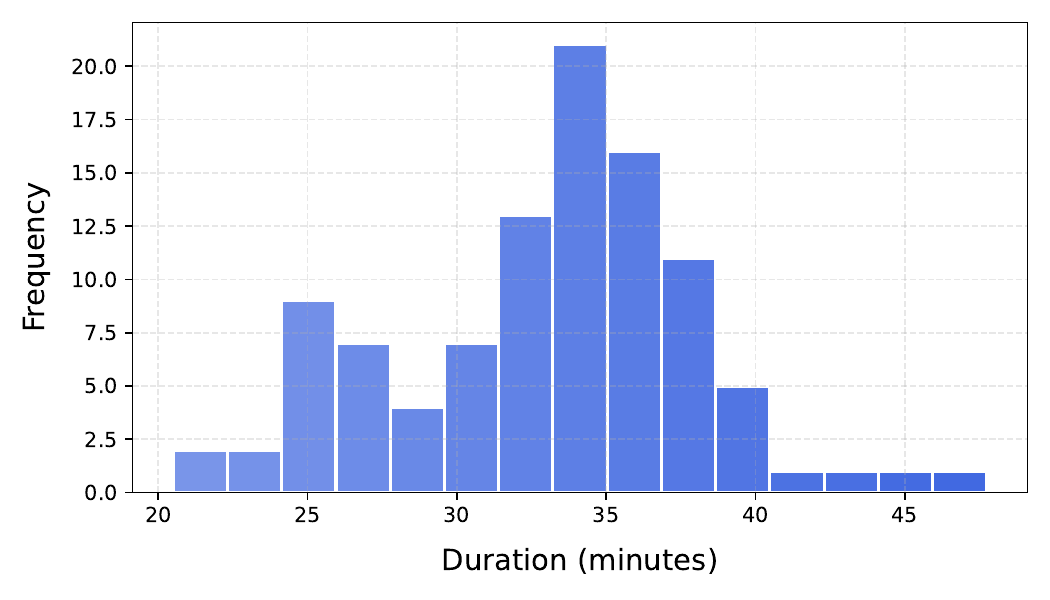}\label{fig:len}}
  \caption{Data analysis: (a) Age-gender structure; (b) Duration distribution.}
  \label{fig:combined}
\end{figure}

\subsection{Statistics}

\begin{wrapfigure}{r}{0.5\columnwidth}
  \vspace{-5mm}
  \includegraphics[width=0.5\columnwidth]{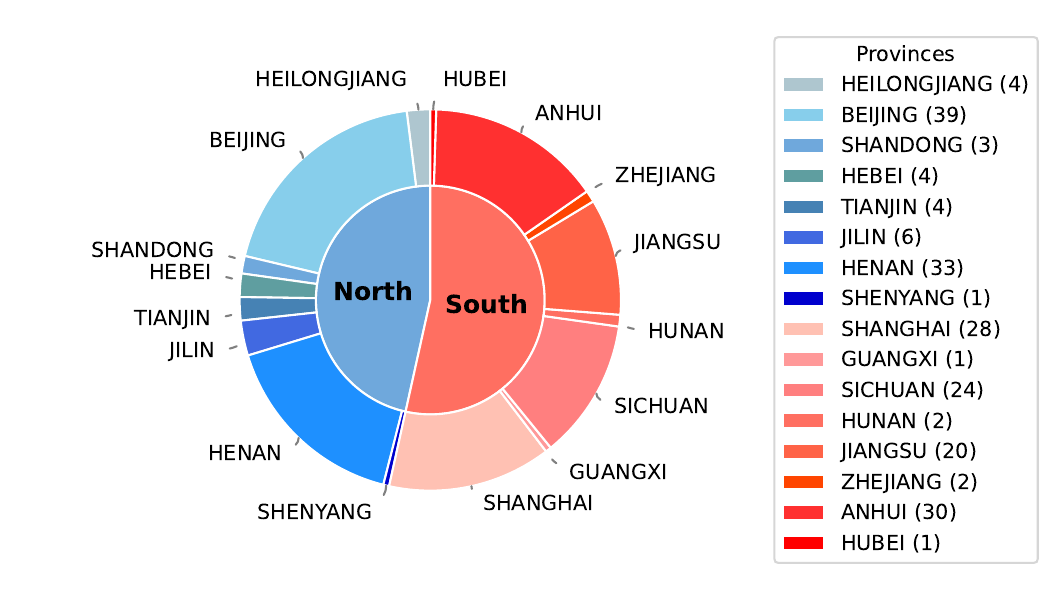}
  \vspace{-8mm}
  \caption{This histogram visualizes the region distribution of the given dataset, showing how the values are spread across different region.}
  \vspace{-4mm}
  \label{fig:regionb}
\end{wrapfigure}

\paragraph{Participants} We recruit 202 participants aged between 75 and 85 years, ensuring a diverse representation across different segments of the ultra-elderly population, and obtain the necessary consent for their participation. Specific authorization details are provided in Appendix~\ref{appendix:authorization}. Figure~\ref{fig:agea} illustrates the gender distribution across different age groups. The dataset includes 67 male and 135 female speakers, with a higher proportion of females. This gender imbalance stems from the relative ease of recruiting female participants during the data collection process, likely due to the higher life expectancy of women in the targeted age range. Geographically, as shown in Figure~\ref{fig:regionb}, 94 participants are from northern China, while 108 are from southern China, covering regions such as Beijing, Shanghai, and Sichuan. This regional diversity enhances dialect recognition models by exposing them to a wide range of linguistic patterns.



\paragraph{Recording} The dataset comprises spontaneous speech dialogues that span a wide range of topics, aimed at capturing the natural flow and diversity of real-life conversations. These dialogues are recorded using various mobile devices, with a distribution of 70\% from Android devices and 30\% from iOS smartphones. The topics of these dialogues are specifically chosen to address relevant issues for older adults, such as health, pets, retirement, and other related matters. Each conversation typically covers between one and three topics, with the distribution of topics visualized in Appendix~\ref{appendix:topic}. Figure~\ref{fig:len} illustrates the distribution of session durations, which range from 25 to 50 minutes, with the majority of sessions clustering around the 35-minute mark.


\paragraph{Annotation} The dataset is annotated by a team of 14 trained annotators who undergo unified training. During the training, they are instructed on a well-defined annotation workflow and strict standards to ensure consistency and accuracy. Detailed guidelines on the annotation process and standards are provided in Appendix~\ref{appendix:annotation}.

Annotations are made across four main dimensions. At the speaker profile level, attributes such as age, gender, and origin are annotated. These annotations are particularly useful for analyzing elderly speech signals, helping to study age-related speech characteristics. The session level includes annotations for temporal segmentation and overlapping speech, which are essential for segmenting speech and identifying overlaps in multi-speaker scenarios. At the utterance level, transcriptions and accent intensity are annotated. Transcriptions support ASR. Finally, at the token level, special sound events like laughter are marked, providing insight into non-verbal communication.

\section{Experiments}
\label{sec:4}
In this section, we assess our dataset across various tasks, including speaker verification, speaker diarization, speech recognition, and speech editing.

\subsection{Speaker Verification}

This section introduces the Speaker Verification (SV) task, which is essential for verifying the identity of speakers within the geriatric population to ensure their financial security. To facilitate this task, we implement a data partitioning strategy, with further details provided in Appendix \ref{appendix:spv_datasplit}.

\subsubsection{Metrics}
We adopt two scoring approaches: probabilistic linear discriminant analysis (PLDA) \citep{prince2007probabilistic} and cosine similarity, with evaluation based on two metrics: (1) \textit{Equal Error Rate (EER)}: We define a threshold $\tau$ where the miss probability equals the false alarm probability. Specifically, if the similarity score is above this threshold, the system accepts that the speakers are the same person; if it is below this threshold, the system rejects the claim. This threshold is selected when the false acceptance rate equals the false rejection rate.  (2) \textit{Minimum Detection Cost (minDCF)}: A cost-sensitive metric for evaluating speaker verification systems under application-specific conditions. 

\subsubsection{Baselines}
We adapt three state-of-the-art speaker embedding systems pre-trained on VoxCeleb \citep{nagrani17_interspeech} through domain-specific fine-tuning via SpeechBrain \citep{speechbrain}:
x-vector architecture \citep{xvector2018},\footnote{\url{https://huggingface.co/speechbrain/spkrec-xvect-voxceleb}} ECAPA-TDNN \citep{ecapa2020},\footnote{\url{https://huggingface.co/speechbrain/spkrec-ecapa-voxceleb}} and ResNet-TDNN \citep{resnet2020}.\footnote{\url{https://huggingface.co/speechbrain/spkrec-resnet-voxceleb}} Detailed hyperparameters are listed in Appendix \ref{appendix:hyperparams-sv}.


\begin{table}[t]
\caption{Results of fine-tuning baselines on the speaker verification task, where Dim indicates the dimension of the extracted embeddings and Dev represents the EER on the validation set.}
\centering
  \resizebox{0.9\textwidth}{!}{
    \begin{tabular}{ccccccccc}
    \toprule
     \multirow{2}{*}{Model} & \multirow{2}{*}{\# Params} & \multirow{2}{*}{Dim} &\multirow{2}{*}{Dev (\%)} & \multicolumn{2}{c}{PLDA} & \multicolumn{2}{c}{Cosine similarity} \\ 
     \cmidrule(lr){5-6} \cmidrule(lr){7-8}
                    &       &                                &                               & EER (\%)        & minDCF       & EER (\%)              & minDCF              \\ 
                           \midrule
                           
    x-vector               &  4.2M & 512                            & 12.04
                           & 17.53            & 0.9032         & 22.38                 & 0.8964                 \\

    ResNet-TDNN           &    15.5M  & 256                            & \textbf{4.37}
                          & 11.39            & 0.7259        & 13.15                & 0.8225 \\

    ECAPA-TDNN         &   20.8M  & 192                            & 5.85
                          & \textbf{8.56  }         & \textbf{0.6263 }        & \textbf{5.66}                 & \textbf{0.6121    }            \\
    \bottomrule
    \end{tabular}
  }

  \label{sv}
\end{table}

\subsubsection{Results and Analysis}

Table \ref{sv} demonstrates two critical findings from our geriatric voice analysis:
First, the strong fintuned models' performance relative to the pretrained model validates the suitability of our dataset for aging voice biometric tasks. This suggests that age-related vocal degradation (e.g., pitch instability, articulatory imprecision) introduces distinct challenges compared to pediatric voices, potentially affecting gender differentiation and speaker discriminability.  The pretrained model results without fine-tuning are provided in Appendix \ref{appendix:sv_baseline_results} for comparison.

Second, despite achieving the best performance on the test set, the ECAPA-TDNN underperforms the ResNet-TDNN on the development set, indicating potential overfitting risks for elderly speech data. This emphasizes the need for data augmentation strategies, regularization techniques and hyperparameter tuning when deploying deep speaker models in geriatric voice applications.

\begin{table}[!t]  
    \caption{Results of fine-tuning speaker embedding extraction models and pretrained model ResNet-34-LM on the speaker diarization task.}  
\centering  
  \resizebox{0.9\textwidth}{!}{  
    \begin{tabular}{ccccccccc}  
    \toprule 
     \multirow{2}{*}{Model} & \multirow{2}{*}{\# Params} & \multirow{2}{*}{Dim} & \multicolumn{2}{c}{collar=0} & \multicolumn{2}{c}{collar=0.25} \\   
     \cmidrule(lr){4-5} \cmidrule(lr){6-7}  
                    &            &                        & DER (\%)        & Confusion (\%)          & DER (\%)            & Confusion (\%)  \\   
    \midrule  

    ResNet-34-LM        &  15.5M & 256                   & 33.14            & 16.82                 & 28.39                & 16.85                \\
          \midrule                   
    x-vector               &  4.2M & 512                    & 53.01            & 36.69                 & 49.82                 & 38.28                 \\

    ResNet-TDNN           &  15.5M & 256                   & 43.44            & 27.13                 & 39.58               & 28.03                \\
    ECAPA-TDNN            &  20.8M & 192                   & \textbf{27.84   }         & \textbf{11.52  }               & \textbf{22.85  }              &\textbf{11.31 }                \\

    \bottomrule  
    \end{tabular}  
  }  

  \label{sd}
\end{table}
\subsection{Speaker Diarization}

This section introduces the speaker diarization task, which entails partitioning audio recordings into segments that correspond to individual speakers. This task is essential for assessing the performance of various speaker models, as detailed in Section \ref{section4.2.3}. The data split is the same as speaker verification task.

\subsubsection{Metrics}
We employ the Diarization Error Rate (DER) as the evaluation metric for the speaker diarization task. The computation of DER is defined as follows:

\begin{equation}
{DER = \frac{FA + MD + Conf}{T},}
\end{equation}

where \textit{\(FA\)} denotes the number of segments incorrectly identified as speech when no speaker is present, \textit{\(MD\)} represents the number of segments where speech is present but not detected. \textit{\(Conf\)} refers to the number of segments where detected speech is attributed to the wrong speaker, and \textit{\(T\)} indicates the total number of speech segments in the reference transcript.

In Table \ref{sd}, we explore two collar settings: 0 seconds and 0.25 seconds. The collar parameter defines a time window around the detected speaker boundaries, allowing for a margin of error in segment alignment. A collar value of 0 seconds requires exact matching of boundaries, while a value of 0.25 seconds introduces a quarter-second tolerance to accommodate minor discrepancies in detection.

\subsubsection{Baselines}
For the speaker diarization task, we use the PyAnnote toolkit \citep{Bredin23, Plaquet23}.\footnote{\url{https://github.com/pyannote/pyannote-audio}} This speaker diarization pipeline consists of three primary components: Voice Activity Detection (VAD), the speaker extractor, and clustering methods (e.g., K-Nearest Neighbors).
\vspace{-2mm}
\begin{itemize}

\item {\bf VAD}: We employ the pyannote/segmentation-3.0 model,\footnote{\url{https://huggingface.co/pyannote/segmentation-3.0}} an end-to-end neural architecture for joint speech activity detection and speaker segmentation. Since false alarm and missed detection rates remain consistent across experimental conditions due to fixed VAD parameters, Table~\ref{sd} exclusively reports confusion errors and DER metrics.

\item {\bf Speaker extractor}: 
For the speaker extractor module, we replace the default ResNet-34-LM extractor from PyAnnote \footnote{\url{https://huggingface.co/pyannote/wespeaker-voxceleb-resnet34-LM}} with fine-tuned versions of x-vector, ECAPA-TDNN, and ResNet-TDNN architectures, all adapted for speaker verification. This yields four distinct experimental configurations, as detailed in Table \ref{sd}.
\item {\bf Clustering method}: 
For the clustering method, we employ the default Spectral Clustering, as proposed by PyAnnote.

\end{itemize}
\vspace{-2mm}

\subsubsection{Results and Analysis}
\label{section4.2.3}

Our experimental analysis yields three principal conclusions regarding elderly speaker diarization:

First, after fine-tuning with the elderly dataset, the ECAPA-TDNN model outperforms the PyAnnote default model ResNet-34-LM by 5.3\% DER improvement at 0 collar and 5.54\% DER imporvement at 0.25 collar. This superiority demonstrates the effectiveness of our dataset for the diarization of elderly speakers.



Second, as shown in Table \ref{sd}, ECAPA-TDNN achieves significantly superior performance over both ResNet-TDNN and x-vector-based baselines on the speaker diarization task. Moreover, consistent with these findings, the ECAPA-TDNN demonstrates consistently lower EER across both PLDA and cosine similarity senarios on the speaker verification task as shown in table \ref{sv}. These results highlight the ECAPA-TDNN’s robustness, confirming its strong generalization ability not only in clean, utterance-level speaker verification tasks but also in more challenging, real-world conversational speaker diarization scenarios involving increased noise and overlapping speech.

Third, our experiments reveal notably high DER and confusion errors. Our analysis attributes this phenomenon to two dataset-specific factors: (1) a pronounced gender imbalance (1:2 female-to-male ratio) contrasting standard benchmarks’ balanced distributions, and (2) age-related vocal changes in elderly speakers that reduce the saliency of secondary sexual voice characteristics. These combined effects create systematic challenges for speaker identity separation, particularly in conversational contexts where demographic diversity and physiological aging patterns naturally occur.

\begin{table}[ht]  
\caption{Decoding performance (CER, \%) of Transformer, Conformer, and E-Branchformer models using Attention rescoring, with accent differentiation and region categorization.}  
\centering  
\begin{tabular}{cccccccccc}  
\toprule  
\multirow{2}{*}{\textbf{Encoder}} & \multirow{2}{*}{\textbf{\# Params}} & \multirow{2}{*}{\textbf{CER}} & \multicolumn{4}{c}{\textbf{Accent}} & \multicolumn{2}{c}{\textbf{Region}} \\
\cmidrule(lr){4-7} \cmidrule(lr){8-9}  
 & & & No  & Light  & Moderate  & Heavy &South & North \\
\midrule  
Transformer  & 14.1M  & 48.99 & 22.58 & 49.05 & 51.07 &80.95& 48.5 & 50.24 \\
Conformer & 15.7M & 34.61 & 21.23 & 34.21 & 37.62 &59.52
  &34.55 & 34.74 \\
E-Branchformer & 16.9M & 33.25 & 20.71 & 33.03 & 35.32 &64.29
& 32.97 & 33.94 \\
\bottomrule
\end{tabular}  

\label{scratch}  
\end{table}

\subsection{Speech Recogition}

Automatic speech recognition entails transcribing spoken language into text, and recognizing elderly speech patterns is crucial in emergency response scenarios due to age-related vocal characteristics that can affect system reliability. This section presents an empirical evaluation of the collected elderly speech corpus, with data partitioning strategies detailed in Appendix \ref{asr_datasplit} .


\subsubsection{Metrics}
The experimental results demonstrate the performance on the test dataset after training with the train dataset, using Character Error Rate (CER) as the evaluation metric, which is computed by the following equation:
\begin{equation}
    CER = \frac{S + D + I}{N},
\end{equation}
where S, D, and I  respectively signify the quantities of substitutions, deletions, and insertions.  denotes the cumulative number of characters within the reference text.
When evaluating character-level transcription accuracy, a system featuring a lower CER is typically regarded as more proficient.

\subsubsection{Baselines}

We employ the open-source Wenet toolkit \citep{wenet} as our training framework, selecting Transformer, Conformer, and E-Branchformer as our baselines. All models are trained using a combined Connectionist Temporal Classification (CTC) and Attention-based Encoder-Decoder (AED) approach. We fine-tune the hyperparameters of these three models to ensure comparable parameter counts while achieving optimal performance. Detailed hyperparameter configurations can be found in the Appendix \ref{appendix:asr-scrach-hyperp} and \ref{appendix:asr-fintune-hyperp}.

The following models are considered in the context of training from scratch: 
\vspace{-2mm}
\begin{itemize}
\item {\bf Transformer}: The standard Transformer architecture employs both CTC and AED objectives, establishing a widely-adopted baseline for ASR.

\item {\bf Conformer}: 
The Conformer \citep{conformer} model integrates convolutions with self-attention for ASR, sandwiched between two feed-forward layers.
\item {\bf E-Branchformer}: Proposed by Kwangyoun Kim et al. \citep{E-Branchformer}, E-Branchformer builds upon the Branchformer \citep{Branchformer}, which attains performance levels comparable to Conformer. E-Branchformer improves on this framework by implementing a novel merging strategy and integrating additional point-wise modules. 
\end{itemize}
\vspace{-2mm}

In addition to the three models trained from scratch, we fine-tune two pre-trained models: Paraformer, which employs the Wenet framework, and Whisper, whose hyperparameters and code base are described in Appendix \ref{appendix:asr-fintune-hyperp}.
\vspace{-2mm}
\begin{itemize}

\item {\bf Paraformer}: Proposed by Gao et al. \citep{paraformer}, Paraformer is a fast and accurate parallel transformer model that leverages a continuous integrate-and-fire (CIF) \citep{cif} predictor to estimate the number of tokens and generate hidden representations. This pre-trained model is trained on a non-public, industry-grade dataset comprising 60,000 hours of Chinese ASR data.

\item {\bf Whisper}: 
Whisper \citep{whisper} \footnote{\url{https://github.com/openai/whisper}} is a Transformer-based multilingual ASR model developed by OpenAI, trained on 680,000 hours of labeled speech data. We examine various versions of Whisper, ranging from tiny to large-v2, with model sizes varying from 39M to 1.55B.
 \vspace{-2mm}

\end{itemize}

\subsubsection{Results and Analysis}

\paragraph{Models Trained from Scratch}

We analyze models trained from scratch across three aspects: baseline performance, accent intensity impact, and regional variations.

\textit{Baseline Model Performance}:  
Table \ref{scratch} compares three models, namely E-Branchformer, Conformer, and Transformer, which are trained from scratch and utilize the attention rescoring decoding method. The E-Branchformer achieves the lowest overall CER, outperforming Conformer by 1.36\% and Transformer by 15.74\%. This performance advantage holds consistently across all tested conditions, including varying accent intensities and geographical regions.

\textit{Accent Intensity Impact}:  
As shown in Table \ref{scratch}, CER increases with accent intensity: Moderate Accent yields higher CER than Light Accent, which in turn exceeds No Accent. This trend highlights the growing recognition challenge as accents become more pronounced. The duration distribution of utterances across different accent levels is presented in Appendix \ref{appendix:accent}.

\textit{Regional Variations}: The E-Branchformer achieves a 0.97\% lower CER in the South than in the North, with the Conformer and Transformer showing improvements of 0.19\% and 1.74\%, respectively. These differences indicate slightly higher recognition difficulty in the North, although overall performance remains comparable. At the provincial level, substantial variations in CER are observed across different provinces. For detailed analysis, please refer to the Appendix\ref{appendix:extra_exp}.

\begin{table}[ht]
\caption{Character Error Rate (CER) (\%) of the pretrained Paraformer-large model, along with various sizes of Whisper models (tiny, base, small, medium, and large-v3) under both zero-shot and fine-tuning settings.}  
\centering
\begin{minipage}{0.65\linewidth}
\centering
\small 
\begin{tabular}{@{}lcccc@{}} 
\toprule  
\textbf{Model}       & \textbf{\# Params} & \textbf{Zero-shot} & \textbf{Fine-tuning} \\
\midrule  
Paraformer-large     & 232M               & 14.91             & 14.41                \\
\midrule
Whisper-tiny         & 39M                & 92.20             & 58.80                    \\
Whisper-base         & 74M                & 64.02             & 38.17                    \\
Whisper-small        & 244M               & 55.83             & 28.69                \\
Whisper-medium       & 769M               & 60.47             & 25.77                \\
Whisper-large-v3     & 1,550M             & 57.74             & 23.84                \\
\bottomrule  
\end{tabular}  
\end{minipage}
\begin{minipage}{0.35\linewidth}

\label{whisper_res}
\end{minipage}
\end{table}

\paragraph{Finetuned Models}

Table \ref{whisper_res} presents comparative CER results for two model families: 1) our Wenet-finetuned Paraformer-large architecture \citep{wenet}, and 2) a series of Whisper models \citep{whisper} adapted through whisper-flamingo fine-tuning.\footnote{\url{https://github.com/roudimit/whisper-flamingo}} The analysis reveals two key findings:

First, although Whisper models show modest CER improvements with larger parameter counts (except for small), all variants have CERs exceeding 50\% in zero-shot senario. This performance gap suggests potential domain mismatch between Whisper's training data distribution and elderly speech characteristics, possibly due to underrepresentation of senior voices in Whisper's pretraining corpus. Such mismatch may explain the frequently observed hallucination patterns in elderly
speech recognition results. Notably, our targeted fine-tuning reduces CERs substantially, improving cer of Whisper-large-v3 from 57.74\% to 23.84\%, demonstrating both the challenge level of our elderly speech dataset and its utility for domain adaptation.

Second, the Paraformer-large model achieves 14.91\% CER in zero-shot evaluation and 14.41\% CER after fine-tuning , outperforming all Whisper variants by significant margins. This advantage likely stems from Paraformer's pretraining on 60,000 hours of proprietary Chinese speech data encompassing diverse regional accents, age groups, and speaking styles. This demonstrates that Paraformer has better generalization ability in the field of Chinese elderly speech recognition.

\begin{table}[ht]   
\caption{Objective Evaluation results of speech editing models trained from scratch.} 
\centering  
\begin{tabular}{@{}lcccc@{}} 
\toprule  
\textbf{Method} &  \textbf{MCD} (↓) & \textbf{STOI} (↑) & \textbf{PESQ} (↑)  \\   
\midrule  
CampNet         & 7.302 & 0.220 & 1.291   \\  
EditSpeech      & 6.225 & 0.514 & 1.363   \\  
A\(^{\mathrm{3}}\)T & 5.851 & 0.586 & 1.455   \\  
FluentSpeech    & 5.811 & 0.627 & 1.645   \\  
\bottomrule  
\end{tabular}  
 
\label{edit}  
\end{table}


\begin{table}[ht]   
\caption{Subjective Evaluation results of speech editing models trained from scratch.} 
\centering  
\begin{tabular}{@{}lccc@{}} 
\toprule  
\textbf{Method} &  \textbf{Boundary-MOS}  & \textbf{MOS}  \\   
\midrule  
CampNet         & 1.25 ± 0.33 & 1.17 ± 0.27   \\  
EditSpeech      & 2.42 ± 0.98 & 1.82 ± 0.91   \\  
A\(^{\mathrm{3}}\)T & 3.00 ± 0.89 & 2.04 ± 0.74    \\  
FluentSpeech    & 4.72 ± 0.44 & 4.53 ± 0.57    \\  
\bottomrule  
\end{tabular}  
 
\label{edit-sub}  
\end{table}

\subsection{Speech Editing}

Speech editing is a generative task that modifies corresponding speech based on text alterations, and editing speech from elderly individuals is particularly useful for editing interviews with seniors. To facilitate this task, we implement a new data split, as detailed in Appendix \ref{appendix:dataspllit-se}. However, utilizing this data for generative tasks poses risks related to elder fraud, particularly in the context of telemarketing scams targeting senior individuals.




\subsubsection{Baselines and Metrics}

We employ the Speech Editing Toolkit \footnote{\url{https://github.com/Zain-Jiang/Speech-Editing-Toolkit}} framework to implement our models, which include CampNet\citep{campnet}, EditSpeech\citep{tan2021editspeech}, A\(^{\mathrm{3}}\)T\citep{a3t} , and FluentSpeech\citep{flu}. The detailed hyperparams are described in Appendix \ref{appenix:se_hper}.

For objective evaluation metrics, we utilize Mel-Cepstral Distortion (MCD) \cite{mcd}, Short-Time Objective Intelligibility (STOI) \cite{stoi}, and Perceptual Evaluation of Speech Quality (PESQ) \cite{pesq}, which are commonly used objective metrics for assessing speech generation quality.

For subjective evaluation metrics, we conducted comprehensive human perceptual evaluation to assess speech editing quality. We randomly selected 50 audio samples and recruited three evaluators to rate each sample on a 1-5 scale using two metrics:

\begin{itemize}
\item {\bf MOS (Mean Opinion Score)}:  Evaluates naturalness of edited speech segments

\item {\bf Boundary-MOS}:  Our specialized metric measuring transition smoothness at editing boundaries

\end{itemize}

\subsubsection{Performance Analysis}
The results presented in Table \ref{edit} and \ref{edit-sub} indicate that FluentSpeech achieved the best performance, with all  objective metrics and subjective metrics falling within acceptable ranges. In relevant research \citep{flu,a3t,campnet,liu2023fluenteditor}, these experimental values are widely recognized as acceptable. This suggests that our dataset is suitable for generative tasks.

\section{Limitations and Ethical Considerations}
\label{sec:5}


While SeniorTalk provides important data for elderly speech processing, it has notable limitations. The dataset includes 202 participants from 16 provinces but displays biases in gender, regional, and accent distributions. Additionally, the participant age range does not fully cover the super-elderly population (85+ years), and the overall dataset size remains relatively modest, which may restrict generalizability across diverse vocal characteristics. From an ethical perspective, rigorous safeguards were implemented: informed consent was obtained, recordings were conducted in controlled environments, and all personally identifiable information was anonymized. Participants received fair compensation in line with local economic conditions. To mitigate risks associated with misuse such as malicious voice synthesis, data access is limited to verified academic researchers under a non-commercial CC BY-NC-SA 4.0 license.



\section{Conclusion}
In this paper, we present SeniorTalk, a Mandarin dataset featuring spontaneous conversations among individuals aged 75 and older. With 55.53 hours of data from 202 speakers across 16 provinces in China, this dataset offers valuable resources for developing voice technologies for aging populations. By providing detailed annotations and conducting extensive experiments across key speech tasks, we establish SeniorTalk as a benchmark for evaluating models for elderly speakers, aiming to bridge the vocal age gap and promote the development of more inclusive voice technologies.

\section*{Acknowledgement}
This work has been supported by the National Key R\&D Program of China (Grant No.2022ZD0116307) and NSFChina (Grant No.62271270).

\bibliographystyle{unsrt}
\bibliography{nips}

\appendix


\newpage
\section{Supplementary Dataset Information}
\label{sec:appendix}

\begin{table}[!htbp]
\caption{Accent Intensity Annotation Guidelines}
\centering
\small
\begin{tabular}{@{}ll@{}}
\toprule
\textbf{Level} & \textbf{Annotation Criteria} \\
\midrule

\multirow{3}{*}{\textbf{No Accent} (0)} 
& \textit{Phonetics}: Fully standard pronunciation with clear articulation \\
& \textit{Comprehensibility}: Effortlessly understood by non-native listeners \\
& \textit{Regional Features}: No detectable regional phonological characteristics \\
\addlinespace

\multirow{3}{*}{\textbf{Light Accent} (1)} 
& \textit{Phonetics}: Occasional non-standard vowels/consonants (<20\% utterances) \\
& \textit{Comprehensibility}: Minor listening effort required for full understanding \\
& \textit{Regional Features}: Subtle but identifiable regional speech patterns \\
\addlinespace

\multirow{3}{*}{\textbf{Moderate Accent} (2)} 
& \textit{Phonetics}: Frequent non-standard prosody/lexical stress (20-50\% utterances) \\
& \textit{Comprehensibility}: Requires focused attention, occasional repetition needed \\
& \textit{Regional Features}: Strong regional phonological markers affecting intelligibility \\
\addlinespace

\multirow{3}{*}{\textbf{Heavy Accent} (3)} 
& \textit{Phonetics}: Pervasive non-standard articulation (>50\% utterances) \\
& \textit{Comprehensibility}: Frequent breakdowns requiring contextual guessing \\
& \textit{Regional Features}: Severely divergent from standard phonological norms \\
\bottomrule
\end{tabular}

\label{appendix:accent_instruction}
\end{table}
\begin{table}[t]
\caption{Annotation Levels and Their Associated Tasks for Dataset Analysis}
\centering
\small
\setlength{\tabcolsep}{6pt} 
\renewcommand{\arraystretch}{1.1} 
\begin{tabularx}{\textwidth}{@{}>{\raggedright}p{2.5cm}X>{\hsize=0.85\hsize}X>{\hsize=1.15\hsize}X@{}}
\toprule
{\raggedright \textbf{Annotation Level}} & 
{\raggedright\textbf{Annotation Dimension}} & 
{\textbf{Associated Tasks}} & 
{\textbf{Representative Instances}} \\
\midrule

\multirow{3}{*}{\raggedright Speaker Metadata} 
& Demographic Age & \multirow{3}{*}{Elderly Speech Analysis } & 75 \\
& Geographic Origin (Province) &  & Jiangsu, Henan \\
& ID Card Gender &  & Female/Male \\ 
\addlinespace
\midrule
\multirow{2}{*}{Session}
& Temporal Segmentation & Speaker Diarization & [48.475 - 73.582] \texttt{spk\_001} \\
& Overlapping Speech & Speech Separation & \texttt{trans1(trans2)[+]} \\
\addlinespace
\midrule
\multirow{2}{*}{Utterance}
& Raw Transcription & Speech Recognition & \texttt{[Mandarin utterance]} \\
& Accent Intensity (0-3) & Ordinal Classification & Neutral (0) / Strong (3) \\
\addlinespace
\midrule
Token 
& Special Markers 
& Paralinguistic Analysis 
& \texttt{[MUSIC],[NOISE],[LAUGHTER]} \\
\bottomrule
\end{tabularx}

\label{hierarchical_annotations}
\end{table}

\subsection{Authorization}
\label{appendix:authorization}

This study employs time-series biometric data collected under a formal ethical authorization framework developed in collaboration with a third-party AI data service provider specializing in biometric acquisition. The framework ensures methodological transparency and regulatory alignment through the following mechanisms:

\textbf{Legally Binding Consent Protocols}
Participants provide informed consent for the collection of vocal and physiological time-series data, explicitly authorizing its use in AI research and derivative model development.

\textbf{Rights Management}
All datasets and derived models remain the exclusive intellectual property of the anonymized provider. Participants retain conditional rights to access, modify, or request data deletion, contingent on technical feasibility (deletion requests that invalidate associated research outputs may require proportional compensation).

\textbf{Cross-Jurisdictional Compliance}
Third-party data sharing requires explicit opt-in consent, with passive approval mechanisms activated only after a 72-hour objection period following notification.

\textbf{Biometric Corpus Specifications}
The dataset includes anonymized temporal features such as age, gender, regional dialect markers, and sequential vocal patterns (e.g., pitch dynamics, spectral entropy trajectories).







\subsection{Topic}
\label{appendix:topic}
\begin{figure}[b]
  \includegraphics[width=1\columnwidth]{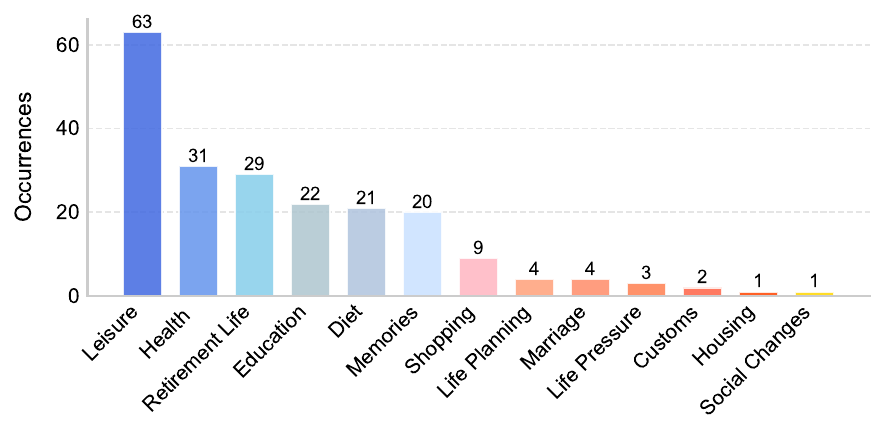}
  \caption{This distribution of topics.}
  \label{fig:topic}
\end{figure}

This section presents the frequency statistics of all topics in the dataset, as illustrated in Figure~\ref{fig:topic}. In total, there are 13 major categories, encompassing 58 distinct topic labels. The distribution of these 13 categories is depicted in the figure. These topics primarily reflect the everyday concerns of elderly individuals, with particular emphasis on areas such as Leisure, Health, and Retirement Life, which appear most frequently. These topics align closely with the key interests and priorities of older adults. Understanding and focusing on these subjects is crucial, as it can greatly benefit the performance of ASR systems, especially when catering to elderly users in relevant contexts.

\subsection{Dataset Accent Distribution Analysis}
\label{appendix:accent}


This section presents the distribution characteristics of accent severity levels within our dataset. The accent annotation criteria follow the standardized framework detailed in table \ref{appendix:accent_instruction}. As illustrated in Table \ref{accent_detail}, the distribution exhibits distinct patterns across different severity levels:

\begin{itemize}
\item Light accent samples constitute the largest temporal proportion (110,419s) and account for 42,790 samples.
\item Moderate accent demonstrates the second-largest distribution with 17,782s duration and 8,226 samples.
\item Samples with no accent (5,951s/8,056) and heavy accent (370s/142) show minor representation.
\end{itemize}

The analysis reveals a pronounced concentration of data in the light accent category, which comprises over half of the total samples, followed by moderate, no, and heavy accents.

\subsection{Annotation}
\label{appendix:annotation}

This section presents information facilitating annotation, covering details about annotators, the annotation process, and the content being annotated.

\begin{table}[b!]
\caption{Annotator Information}
\centering
\begin{tabular}{lcccc}
\toprule
ID & Region & Age & Gender & Education \\
\midrule
1 & Henan & 36 & Male & Bachelor \\
2 & Hebei & 23 & Female & Bachelor \\
3 & Hebei & 24 & Female & Bachelor \\
4 & Hebei & 24 & Female & Bachelor \\
5 & Hebei & 30 & Female & Bachelor \\
6 & Fujian & 27 & Male & Bachelor \\
7 & Fujian & 23 & Male & Bachelor \\
8 & Fujian & 23 & Female & Bachelor \\
9 & Fujian & 38 & Female & Bachelor \\
10 & Chuzhou & 22 & Male & Bachelor \\
11 & Chuzhou & 27 & Male & Bachelor \\
12 & Chuzhou & 23 & Female & Bachelor \\
13 & Yunnan & 27 & Male & Bachelor \\
14 & Hunan & 47 & Male & Bachelor \\
\bottomrule
\end{tabular}

\label{tab:annotator_demographics}
\end{table}

\begin{table}[t]
\caption{Accent distribution}
\centering
\begin{tabular}{lcc}
\toprule
Accent & Total Duration (seconds) & Number of Samples \\
\midrule
No       &   5,951 &  8,056 \\
Light    & 110,419 & 42,790 \\
Moderate &  17,782 &  8,226 \\
Heavy    &    370 &   142 \\
\bottomrule
\end{tabular}
\label{accent_detail}
\end{table}



\subsubsection{Annotator}

As shown in Table~\ref{tab:annotator_demographics}, We have a total of 14 annotators. Eleven of them are between 23 and 30 years old, and three are between 30 and 50 years old. All annotators hold undergraduate degrees. Moreover, they are proficient in their local dialects, which facilitates the transcription of accented speech.

\subsubsection{Annotation workflow}

The annotation process utilizes a proprietary cloud-based platform developed by a third-party data service provider. Annotators access tasks through this platform, which facilitates manual segmentation of conversational data into discrete utterances followed by automated speech recognition to generate preliminary transcripts. These baseline outputs undergo human revision alongside annotation of linguistic features including accent intensity levels (see Table~\ref{appendix:accent_instruction}) and paralinguistic markers such as [NOISE] or [MUSIC]. A multi-tier quality assurance protocol is implemented through the same platform, where dedicated reviewers systematically validate annotation consistency and accuracy prior to dataset finalization.  


\subsubsection{Annotation Information}


Table~\ref{hierarchical_annotations} presents our four-tier annotation framework comprising speaker, session, utterance, and token levels, with each layer supporting distinct speech processing tasks through systematically designed metadata. 

The speaker-level annotations capture demographic attributes (geographic origin, gender, age) to enable geriatric voice analysis. 

At the session level, comprehensive temporal annotations of turn-taking boundaries and overlap detection facilitate speaker diarization and speech separation benchmarking. 

Utterance-level transcriptions incorporate accent intensity labels and orthographic normalization, jointly supporting automatic speech recognition and ordinal classification of vocal aging patterns. 

The token tier extends this granularity with paralinguistic markers, including [MUSIC], [NOISE], [LAUGHTER], and [SONANT], enabling quantitative analysis of non-lexical vocal characteristics critical for elderly communication studies.

\subsection{Licenses for assets}
\label{app:licenses}
This work adheres to strict licensing compliance for all third-party assets. Below are the licensing terms for codebases and datasets utilized in our speech dataset creation paper:
\begin{itemize}
    \item \textbf{WeNet:} The WeNet framework \cite{wenet} for end-to-end speech recognition is licensed under Apache 2.0, permitting modification and commercial use with attribution. 
    
    \item \textbf{Whisper Flamingo:} The Whisper Flamingo framework \cite{whisper} for Whisper adaptation operates under BSD 3-Clause License, requiring copyright retention in derivative works.
    
    \item \textbf{SpeechBrain:} SpeechBrain toolkit \cite{speechbrain} utilizes Apache 2.0. 

    \item \textbf{PyAnnote:} PyAnnote toolkit\citep{Bredin23, Plaquet23} utilizes MIT License which allows users to freely use, modify, and distribute the software, provided that the copyright notice and permission notice are retained. 

    \item \textbf{Speech-Editing-Toolkit:} The Speech-Editing-Toolkit \cite{flu} includes no explicit software license but enforces strict usage prohibitions through appended declaration:

\begin{quote}
\small
``Any synthesis of recognizable voices (e.g., politicians, celebrities) without explicit consent violates ethical standards and copyright laws. Violators assume full legal liability.''  
\end{quote}

    \item \textbf{VoxCeleb Dataset:} Training data for speaker modeling is governed by CC BY 4.0 \cite{nagrani17_interspeech}, mandating proper attribution for academic/commercial redistribution.
\end{itemize}

For reproducibility and open science, we release all components of our SeniorTalk under the following licenses:

\begin{itemize}
    \item \textbf{SeniorTalk Dataset:} Our conversational speech dataset is released under CC BY-NC-SA 4.0, restricting commercial applications while requiring share-alike terms for derivatives.
    
    \item \textbf{Codebase:} Core infrastructure follows Apache License 2.0. Embedded third-party code retains original licensing.
\end{itemize}

\section{Supplementary Experiments}
\label{appendix:Supplementary_Experiments}
This section mainly presents supplementary experimental details, including the dataset partitioning for each task and the hyperparameters. The fine-tuning and zero-shot tests of the Whisper model are performed using an NVIDIA A800, while all other experiments are carried out with an NVIDIA GeForce RTX 3090.
\subsection{Data Split}
\label{appendix:B}

\begin{table}[htbp]
\caption{Summary of dataset splits, including the number of speakers (\# Spk.) and utterances (\# Utt.), total duration (Dur.), and average utterance length (Avg.) for speech recognition task.}
\centering
\begin{tabular}{ccccc}
\toprule
Split & \# Spk. & \# Utt. & Dur. (hrs) & Avg. (s) \\ 
\midrule
Train & 162     & 47,269     & 29.95            & 2.28     \\
Dev   & 20      & 6,891      & 4.09             & 2.14     \\
Test  & 20      & 5,869      & 3.77             & 2.31     \\
Sum   & 202     & 60,029     & 37.81            & 2.27    \\ 
\bottomrule
\end{tabular}

\label{asr_dataset_detail2}
\end{table}
\begin{table}[!ht]
\caption{Summary of dataset splits, including the speakers range (\# Spk.), utterances (\# Utt.), total duration (Dur.), and average utterance length (Avg.) for speaker verification and speaker dirazation tasks.}
\centering
\begin{tabular}{ccccc}
\toprule
Split & \# Spk. & \# Utt. & Dur. (hrs) & Avg. (s) \\ \midrule
Train & 0-182      & 48,591     &  30.47            & 2.26     \\
Dev   &  0-182    & 5,398    & 3.40             & 2.27     \\
Test  & 182-202   & 6,040       & 3.95             & 2.35    \\
Sum   & 0-202     & 60,029     & 37.82          & 2.27    \\ \bottomrule
\end{tabular}

\label{dataset_split_verification}
\end{table}

\subsubsection{Speaker Verification and Diarization}
\label{appendix:spv_datasplit}

For the speaker verification experiments, we first segment the original dialogue dataset into sentence-level units. Given that the dataset includes precise timestamps for each sentence and specific audio markers for events such as overlapping speech and noise, we extract clean audio segments devoid of these special sound events and speaker overlaps. From the 101 dialogues in the dataset, we randomly select 10 dialogues, using the corresponding segmented audio at the sentence level as our test dataset. The remaining audio is split, with 10\% randomly allocated for the validation set and 90\% designated for the training set. The detailed information of the datasplit is shown in table \ref{dataset_split_verification}. After the datasplit, we create 20,000 carefully balanced verification pairs (50\% genuine vs. 50\% impostor pairs) from the test set. This trial composition ensures uniform coverage of both intra-speaker 
${(spk_i, spk_j)}$ and inter-speaker pairs ${(spk_i, spk_j)}$. The detailed datasplit information is shown in table \ref{dataset_split_verification}. For speaker diarization, the dataset split is identical to that of the speaker verification task, except that we use dialogues instead of sentences.

\begin{table}[htbp]
\caption{Summary of dataset splits, including the number of speakers (\# Spk.) and utterances (\# Utt.), total duration (Dur.), and average utterance length (Avg.) for speech recognition task.}
\centering
\begin{tabular}{ccccc}
\toprule
Split & \# Spk. & \# Utt. & Dur. (hrs) & Avg. (s) \\ 
\midrule
Train & 162     & 47,269     & 29.95            & 2.28     \\
Dev   & 20      & 6,891      & 4.09             & 2.14     \\
Test  & 20      & 5,869      & 3.77             & 2.31     \\
Sum   & 202     & 60,029     & 37.81            & 2.27    \\ 
\bottomrule
\end{tabular}

\label{asr_dataset_detail2}
\end{table}

\subsubsection{Speech Recognition}
\label{asr_datasplit}

For the elderly dataset, we initially segment the dialogue-based data into clean sentence-level data following the annotations. This process is similar to the data split in the speaker verification task described in Appendix \ref{appendix:spv_datasplit}. Subsequently, we partition the data by speaker and randomly divide it into three subsets: the training set, the development set (dev), and the test set, with a ratio of 8:1:1. The detailed data split information is presented in Table \ref{asr_dataset_detail2}.

\subsubsection{Speech Editing}
\label{appendix:dataspllit-se}
After converting the conversation into sentences using the method described in the speaker verification data split in Appendix \ref{appendix:spv_datasplit}, we filter the sentences. Only those with more than four characters are included for training and testing. We then divide the data into a training-to-testing ratio of 9:1.
\begin{table}
\caption{Summary of dataset splits, including the  utterances (\# Utt.), total duration (Dur.), and average utterance length (Avg.) for speech editing tasks.}
\centering
\begin{tabular}{cccc}
\toprule
Split & \# Utt. & Dur. (hrs) & Avg. (s) \\ \midrule
Train &    44,954     & 32.42            & 2.60    \\
Test  &    4,994      & 3.68            & 2.65     \\
Sum   &    49,948     & 36.10            & 2.60    \\ \bottomrule
\end{tabular}

\label{dataset_detail}
\end{table}

\subsection{Hyperparams}
\label{appendix:hyperparams}

\begin{table}[htbp]
 \caption{Hyperparameters for training ASR models from scratch.}
    \centering
    \begin{tabular}{lccccc}
        \toprule
        Encoder &  Accum Grad & Batch size & LR & Warmup & Epochs \\
        \midrule
        Transformer & 4 & 4& 1e-3 & 5,000 & 100 \\
        Conformer  & 4 & 4& 1e-3 & 5,000 & 60 \\
        E-Branchformer  & 4 & 4 & 5e-4 & 5,000 & 60 \\
        \bottomrule
    \end{tabular}
   
    \label{tab:A.1}
\end{table}

\begin{table}[htbp]
\caption{Hyperparameters for fine-tuning pre-trained ASR models.}
    \centering
    
    \begin{tabular}{lccccc}
        \toprule
        Model &   Accum Grad & Batch size & Learning rate & Warmup & Training steps \\
        \midrule
        Paraformer-large & 4& 28 & 5e-4& 5,000 & 101,579 \\
        Whisper-large & 1 & 8 & 5e-6 & 1,000 & 90,000 \\
        
        Whisper-medium & 1 & 3 & 6.25e-6 & 1,000 & 225,000 \\
        Whisper-small & 1 & 4 & 1.25e-5 & 1,000 & 225,000 \\
        \bottomrule

    \end{tabular}
    
    \label{tab:A.2}
\end{table}
\begin{table}[htbp]
        \caption{Hyperparameters for training speaker verification models.}
    \centering
    \begin{tabular}{lccccc}
        \toprule
        Model & Batch size & LR Schedule & Init LR & Base LR & Epochs \\
        \midrule
        ECAPA-TDNN & 128 & Cyclic & 5e-3 & 1e-8 & 20 \\
        ResNet-TDNN & 128 & Cyclic & 5e-3 & 1e-8 & 20 \\
        x-vector & 128 & Linear & 5e-3 & 1e-4 & 20 \\
        \bottomrule
    \end{tabular}

    \label{tab:A.3}
\end{table}

\begin{table}[htbp]
 \caption{Hyperparameters for training speaker editing models.}
    \centering

    \begin{tabular}{lcccccc}
        \toprule
        Model &   Params &Batch size & LR & Warmup & Training steps \\
\midrule  

CampNet  & 21.22M & 16 & 2e-4 & 8,000 & 200,000   \\

EditSpeech &48.15M & 16 & 2e-4 & 8,000 & 200,000  \\

A\(^{\mathrm{3}}\)T  & 14.86M & 16 & 2e-4 & 8,000& 200,000   \\

FluentSpeech &23.86M & 30 & 2e-4 & 8,000 & 200,000   \\

\bottomrule  
    \end{tabular}
       
    \label{tab:A.4}
\end{table}

\subsubsection{ASR Model Training from Scratch}
\label{appendix:asr-scrach-hyperp}
The hyperparameters used for training ASR models from scratch are presented in Table \ref{tab:A.1}.  We experimented with three different encoder architectures: Transformer, Conformer, and E-Branchformer. For all models, we used an accumulation gradient of 4 and a batch size of 4.  The initial learning rate was set to 1e-3 for Transformer and Conformer models, while it was 5e-4 for the E-Branchformer model.  A warmup period of 5,000 steps was used for all models. The models were trained for a maximum of 100 epochs (Transformer) and 60 epochs (Conformer and E-Branchformer).

\subsubsection{ASR Model Fine-tuning}
\label{appendix:asr-fintune-hyperp}
Table \ref{tab:A.2} outlines the hyperparameters used for fine-tuning pre-trained ASR models. Our experiments covered Paraformer-large and multiple Whisper variants (large/medium/small). The batch size varied depending on the model, ranging from 3 to 28.  The learning rates were carefully chosen for each model, spanning from 5e-6 to 5e-4.  A warmup period of 1,000 steps was used across all fine-tuned models. The total training steps also varied significantly depending on the model, reflecting the different sizes and pre-training strategies. All the hyperparameters of whisper models are the default setting of whisper-flamingo.

\subsubsection{Speaker Verification Model Training}
\label{appendix:hyperparams-sv}
Training details for our speaker verification models, including ECAPA-TDNN, ResNet-TDNN and x-vector architectures, are summarized in Table \ref{tab:A.3}. All models used a batch size of 128.  For ECAPA-TDNN and ResNet-TDNN, a cyclic learning rate schedule was employed with an initial learning rate of 5e-3 and a base learning rate of 1e-8.  The x-vector model utilized a linear learning rate schedule with an initial learning rate of 5e-3 and a base learning rate of 1e-4.  All models were trained for 20 epochs.

\begin{table}[ht]
\caption{Model Performance in Southern Provinces}
    \centering

    \begin{tabular}{l|ccccccc}
        \toprule
        Model & Shanghai & Guangxi & Sichuan & Anhui & Zhejiang & Jiangsu & Hunan\\
        \midrule
        Paraformer-fintuned & 13.68 & 4.88 & 9.62 & 16.11 & 12.42 & 16.57 & 14.60 \\
        Paraformer & 13.26 & 6.24 & 10.07 & 17.91 & 12.30 & 16.61 & 13.18 \\
        Brachformer & 27.37 & 10.82 & 25.94 & 36.58 & 29.83 & 36.68 & 39.94 \\
        Conformer & 29.11 & 11.91 & 27.11 & 38.01 & 30.17 & 39.44 & 40.65 \\
        Tranformer & 42.27 & 26.91 & 41.78 & 52.3 & 46.53 & 51.82 & 54.90 \\
        \bottomrule
    \end{tabular}
        
    \label{tab:southern_model_performance}

\end{table}
\begin{table}[!t]
\caption{Model Performance in Northern Provinces}
\centering

    \begin{tabular}{l|ccccc}
        \toprule
        Model & Beijing & Heilongjiang & Liaoning & Hebei & Henan\\
        \midrule
        Paraformer-fintuned & 7.38 & 5.85 & 27.08 & 7.71 & 20.93 \\
        Paraformer & 6.80 & 5.38 & 23.85 & 9.41 & 24.19 \\
        Brachformer & 23.65 & 21.21 & 55.31 & 19.63 & 40.85 \\
        Conformer & 25.15 & 20.27 & 56.88 & 20.44 & 41.59 \\
        Tranformer & 43.01 & 39.61 & 67.58 & 36.61 & 56.03 \\
        \bottomrule
    \end{tabular}
        
    \label{tab:northern_model_performance}

\end{table}

\subsubsection{Speech Editing Model Training}
\label{appenix:se_hper}

For training our second set of Speech Editing models(CampNet, EditSpeech, A3T, and FluentSpeech), we utilized the hyperparameters summarized in Table \ref{tab:A.4}. The number of trainable parameters for each model is also provided.  All models, except FluentSpeech, used a batch size of 16. FluentSpeech used a batch size of 30.  The initial learning rate was set to 2e-4 for all models.  A warmup period of 8,000 steps was used, and all models were trained for 200,000 steps.

\subsection{Extra Experiment Analysis}

\begin{table}
\centering
\caption{Paraformer Model Performance across age}
\begin{tabular}{lrrrrrrrr}  
\toprule
model & age & S & D & I & N & wer(\%) & num\_utterances & num\_speakers \\
\midrule
Paraformer & 75 & 1020 & 217 & 195 & 14452 & 9.91 & 1278 & 5 \\
& 76 & 2008 & 405 & 451 & 21247 & 13.48 & 2072 & 7 \\
& 78 & 1233 & 263 & 176 & 8546 & 19.56 & 1010 & 3 \\
& 80 & 769 & 102 & 65 & 5430 & 17.24 & 602 & 2 \\
& 81 & 918 & 153 & 153 & 6331 & 19.33 & 621 & 2 \\
& 82 & 489 & 51 & 91 & 2761 & 22.85 & 286 & 1 \\
\bottomrule
\label{tab:age_wer1}
\end{tabular}
\end{table}

\begin{table}
\centering
\caption{Paraformer-finetuned Model Performance}
\begin{tabular}{lrrrrrrrr}  
\toprule
model & age & S & D & I & N & wer(\%) & num\_utterances & num\_speakers \\
\midrule
Paraformer-finetuned & 75 & 906 & 268 & 231 & 14452 & 9.72 & 1278 & 5 \\
& 76 & 1994 & 520 & 496 & 21247 & 14.17 & 2072 & 7 \\
& 78 & 986 & 288 & 223 & 8546 & 17.52 & 1010 & 3 \\
& 80 & 734 & 107 & 85 & 5430 & 17.05 & 602 & 2 \\
& 81 & 749 & 167 & 170 & 6331 & 17.15 & 621 & 2 \\
& 82 & 407 & 42 & 91 & 2761 & 19.56 & 286 & 1 \\
\bottomrule
\label{tab:age_wer2}
\end{tabular}
\end{table}

\subsubsection{Detailed results of Wenet models on ASR task}
\label{appendix:extra_exp}
This section primarily presents the detailed CER of the models trained with Wenet in the speech recognition experiments. Specifically, the specific recognition results of these models for sentences from different provinces are shown in Table~\ref{tab:northern_model_performance} and Table~\ref{tab:southern_model_performance}. In Table \ref{tab:northern_model_performance}, we can observe that the CERs of Beijing, Hebei, and Heilongjiang are relatively low, while the recognition difficulty for Liaoning and Henan increases. Similarly, in Table \ref{tab:southern_model_performance}, Guangxi has the lowest recognition difficulty among all provinces. Moreover, the order of recognition difficulty, that is, the ranking of CERs, among other provinces is roughly the same. This indicates that when we break down the regions to the provincial level, the recognition difficulty varies across different provinces.

We have conducted further analysis on age effects. Table \ref{tab:age_wer1} and \ref{tab:age_wer2} are our experimental results for the Paraformer model before and after fine-tuning. Our analysis indicates that the difficulty of speech recognition increases significantly with speaker age, which underscores the necessity of our dataset in addressing this challenge.

\begin{table}[!t]
\centering
\caption{Results of baselines on the speaker verification task}
\label{tab:eer_comparison}
\begin{tabular}{lcc} 
\toprule
\textbf{Model} & \textbf{PLDA EER(\%)} & \textbf{Cosine EER(\%)} \\
\midrule
x-vector       & 18.13               & 30.81               \\
ResNet-TDNN    & 15.81               & \textbf{17.50}       \\
ECAPA-TDNN     & \textbf{11.40}      & 20.88               \\
\bottomrule
\end{tabular}
\end{table}

\subsubsection{Results of baselines on the speaker verification task}
\label{appendix:sv_baseline_results}


This section evaluates three speaker verification models trained exclusively on the VoxCeleb dataset without SeniorTalk-based fine-tuning, as summarized in Table \ref{tab:eer_comparison}. After applying our proposed fine-tuning strategy using the SeniorTalk dataset, the models exhibit significant performance improvements on elderly speaker verification tasks, highlighting the critical role of domain-specific adaptation for aging voice biometrics.

\section{Ethics Statement}
\label{sppendix:ethics}

This study strictly adhered to rigorous ethical protocols to safeguard elderly participants' rights. Audio recordings were conducted in quiet indoor environments at senior care facilities. To accommodate potential cognitive decline among participants, multiple conversation topics were provided during dyadic interactions, with recording devices positioned equidistant between paired participants.

Prior to each session, informed consent was obtained after explaining the research objectives and data collection parameters, including voice characteristics, conversational content, age documentation, and accent analysis. Participants were compensated monetarily, with amounts ranging from 330 to 400 RMB. This compensation was calibrated according to local purchasing power, thereby ensuring equitable remuneration tailored to the specific geographical locations of the participants. All personal identifiers (e.g., national ID numbers, full names) were systematically de-identified by replacing them with unique speaker identifiers, despite initial age verification requiring temporary ID inspection.

The dataset carries inherent risks requiring stringent governance. Potential malicious exploitation for voice synthesis could potentially exacerbate elderly-targeted telecommunications fraud through voice spoofing. Accordingly, access is strictly restricted to vetted academic researchers through institutional credential verification, with legally binding agreements prohibiting commercial use or redistribution.

Key ethical safeguards implemented include: 1) Explicit participant consent through verbal and written confirmation, 2) Comprehensive privacy preservation via anonymization protocols, 3) Fair compensation aligned with regional economic standards, 4) Institutional Review Board approval for all data collection procedures, 5) Multi-layered access controls preventing unauthorized usage.

This framework ensures compliance with international research ethics standards while balancing scientific utility with participant protection in vulnerable populations.


\newpage
\section*{NeurIPS Paper Checklist}

\begin{enumerate}

\item {\bf Claims}
    \item[] Question: Do the main claims made in the abstract and introduction accurately reflect the paper's contributions and scope?
    \item[] Answer: \answerYes{} 
    \item[] Justification: The abstract and introduction accurately reflect the paper’s contributions and scope.

\item {\bf Limitations}
    \item[] Question: Does the paper discuss the limitations of the work performed by the authors?
    \item[] Answer: \answerYes{} 
    \item[] Justification: We summarise limitations in
Section  \ref{sec:5}.

\item {\bf Theory assumptions and proofs}
    \item[] Question: For each theoretical result, does the paper provide the full set of assumptions and a complete (and correct) proof?
    \item[] Answer: \answerNA{} 
    \item[] Justification: 

    \item {\bf Experimental result reproducibility}
    \item[] Question: Does the paper fully disclose all the information needed to reproduce the main experimental results of the paper to the extent that it affects the main claims and/or conclusions of the paper (regardless of whether the code and data are provided or not)?
    \item[] Answer: \answerYes{} 
    \item[] Justification: We publicly release the source code on GitHub (\url{https://github.com/flageval-baai/SeniorTalk}). In Section \ref{sec:4}, we comprehensively introduce the models employed in our study along with the associated repositories. Subsequently, in Appendix \ref{appendix:hyperparams}, we delve into a detailed discussion of the hyperparameter settings utilized in our experiments. Moreover, Appendix \ref{appendix:B} is dedicated to an exploration of the dataset partitioning strategies adopted for our experimental evaluations.

\item {\bf Open access to data and code}
    \item[] Question: Does the paper provide open access to the data and code, with sufficient instructions to faithfully reproduce the main experimental results, as described in supplemental material?
    \item[] Answer: \answerYes{} 
    \item[] Justification: In abstract, we release the code and data link. In Section \ref{sec:4}, we introduce the baseline models along with the associated repositories. Subsequently, in Appendix \ref{appendix:hyperparams}, we delve into a detailed discussion of the hyperparameter settings utilized in our experiments. Moreover, Appendix \ref{appendix:B} is dedicated to an exploration of the dataset partitioning strategies adopted for our experimental evaluations.

\item {\bf Experimental setting/details}
    \item[] Question: Does the paper specify all the training and test details (e.g., data splits, hyperparameters, how they were chosen, type of optimizer, etc.) necessary to understand the results?
    \item[] Answer: \answerYes{} 
    \item[] Justification: in Appendix \ref{appendix:hyperparams}.

\item {\bf Experiment statistical significance}
    \item[] Question: Does the paper report error bars suitably and correctly defined or other appropriate information about the statistical significance of the experiments?
    \item[] Answer: \answerNo{} 
    \item[] Justification: Error bars are not reported in our study due to the substantial computational expense involved, particularly given the large number of experimental spacecraft we have utilized. Incorporating error bars would necessitate considerable additional time and resources, which are not feasible within the scope of our current research constraints.

\item {\bf Experiments compute resources}
    \item[] Question: For each experiment, does the paper provide sufficient information on the computer resources (type of compute workers, memory, time of execution) needed to reproduce the experiments?
    \item[] Answer: \answerYes{} 
    \item[] Justification: In Supplementary Experiments \ref{appendix:Supplementary_Experiments}, we conduct a detailed exploration of the experimental setup, including the hardware configuration of the machines utilized and the specific training steps.
    
\item {\bf Code of ethics}
    \item[] Question: Does the research conducted in the paper conform, in every respect, with the NeurIPS Code of Ethics \url{https://neurips.cc/public/EthicsGuidelines}?
    \item[] Answer: \answerYes{} 
    \item[] Justification: In Section \ref{sec:5}, we provide a concise overview of the topic, while a comprehensive discussion is presented in Appendix \ref{sppendix:ethics}.

\item {\bf Broader impacts}
    \item[] Question: Does the paper discuss both potential positive societal impacts and negative societal impacts of the work performed?
    \item[] Answer: \answerYes{} 
    \item[] Justification: in Appendix \ref{sppendix:ethics}.
    
\item {\bf Safeguards}
    \item[] Question: Does the paper describe safeguards that have been put in place for responsible release of data or models that have a high risk for misuse (e.g., pretrained language models, image generators, or scraped datasets)?
    \item[] Answer: \answerYes{} 
    \item[] Justification: We discuss the safeguards in Appendix \ref{sppendix:ethics} and all the data and code are released under the CC BY-NC-SA 4.0 license.

\item {\bf Licenses for existing assets}
    \item[] Question: Are the creators or original owners of assets (e.g., code, data, models), used in the paper, properly credited and are the license and terms of use explicitly mentioned and properly respected?
    \item[] Answer: \answerYes{} 
    \item[] Justification: in appendix \ref{app:licenses}

\item {\bf New assets}
    \item[] Question: Are new assets introduced in the paper well documented and is the documentation provided alongside the assets?
    \item[] Answer: \answerYes{} 
    
    \item[] Justification: Compliance documentation for voice data collection ethics (including participant consent protocols) and licensing specifications are detailed in Appendix~\ref{sec:appendix}. Implementation specifics, including hyperparameter configurations and training procedures, are provided in Supplementary Materials (Appendix~\ref{appendix:Supplementary_Experiments}).

\item {\bf Crowdsourcing and research with human subjects}
    \item[] Question: For crowdsourcing experiments and research with human subjects, does the paper include the full text of instructions given to participants and screenshots, if applicable, as well as details about compensation (if any)? 
    \item[] Answer: \answerYes{}{} 
    \item[] Justification: in appendix \ref{sec:appendix} and appendix \ref{sppendix:ethics}

\item {\bf Institutional review board (IRB) approvals or equivalent for research with human subjects}
    \item[] Question: Does the paper describe potential risks incurred by study participants, whether such risks were disclosed to the subjects, and whether Institutional Review Board (IRB) approvals (or an equivalent approval/review based on the requirements of your country or institution) were obtained?
    \item[] Answer: \answerNo{} 
    \item[] Justification: While formal IRB approval was not obtained for this study, all data collection strictly followed ethical research practices including explicit participant consent, voice data anonymization, and compliance with international privacy regulations (GDPR/CCPA). We implemented robust safeguards against potential risks while adhering to the NeurIPS Code of Ethics throughout the research process.

\item {\bf Declaration of LLM usage}
    \item[] Question: Does the paper describe the usage of LLMs if it is an important, original, or non-standard component of the core methods in this research? Note that if the LLM is used only for writing, editing, or formatting purposes and does not impact the core methodology, scientific rigorousness, or originality of the research, declaration is not required.
    \item[] Answer: \answerNA{} 
    \item[] Justification: We do not use the LLM.

\end{enumerate}

\end{document}